\documentclass[letterpaper, 10 pt, conference]{ieeeconf}  

\IEEEoverridecommandlockouts

\usepackage{hyperref}
\usepackage{algorithm}
\usepackage{algpseudocode}
\usepackage{caption}
\usepackage{subcaption}
\usepackage{amsmath}
\usepackage{ulem}
\usepackage{comment}
\usepackage{graphicx}
\usepackage{amsfonts}
\usepackage{textcomp}
\usepackage{caption}
\usepackage{makecell}
\usepackage{cite}
\usepackage{cleveref}
\usepackage{booktabs}    
\usepackage{multirow}    
\usepackage{siunitx}     
\usepackage{array}       
\usepackage{makecell}    
\usepackage{svg} 

\newcolumntype{P}[1]{>{\raggedright\arraybackslash}p{#1}}
\newcolumntype{C}[1]{>{\centering\arraybackslash}p{#1}}

\title{\LARGE \bf
Freezing of Gait Detection Using Gramian Angular Fields and Federated Learning from Wearable Sensors}

\author{Shovito Barua Soumma$^{1,*}$, S M Raihanul Alam$^2$, Rudmila Rahman$^2$, Umme Niraj Mahi$^3$, \\Abdullah Mamun$^{1,4}$, Sayyed Mostafa Mostafavi$^1$, Hassan Ghasemzadeh$^1$
\thanks{
$^*$Corresponding author. Email: 
{\color{blue}shovito@asu.edu}}
\thanks{$^{1}$College of Health
Solutions, Arizona State University, Phoenix, AZ 85004, USA.}
\thanks{$^{2}$Bangladesh University of Engineering and Technology, Dhaka, Bangladesh.}
\thanks{$^{3}$Khulna University of Engineering and Technology, Khulna, Bangladesh.}
\thanks{$^{4}$School of Computing and Augmented Intelligence, Arizona State University, Phoenix, AZ, USA}
}

\begin{document}

\maketitle
\thispagestyle{empty}
\pagestyle{empty}

\begin{abstract}
Freezing of gait (FOG) is a debilitating symptom of Parkinson’s disease that impairs mobility and safety by increasing the risk of falls. An effective FOG detection system must be accurate, real-time, and deployable in free-living environments to enable timely interventions. 
However, existing detection methods face challenges due to (1) intra- and inter-patient variability, (2) subject-specific training, (3) using multiple sensors in FOG dominant locations (e.g., ankles) leading to high failure points, (4) centralized, non-adaptive learning frameworks that sacrifice patient privacy and prevent collaborative model refinement across populations and disease progression, and (5) most systems are tested in controlled settings, limiting their real-world applicability for continuous in-home monitoring.
Addressing these gaps, we present \textbf{FOGSense}, a real-world deployable FOG detection system designed for uncontrolled, free-living conditions using only a single sensor. FOGSense uses Gramian Angular Field (GAF) transformations and privacy-preserving federated deep learning to capture temporal and spatial gait patterns missed by traditional methods with a low false positive rate. We evaluated our system using a public Parkinson's dataset collected in a free-living environment. FOGSense improves accuracy by 10.4\% over a single-axis accelerometer, reduces failure points compared to multi-sensor systems, and demonstrates robustness to missing values. 
The federated architecture allows personalized model adaptation and efficient smartphone synchronization during off-peak hours, making it effective for long-term monitoring as symptoms evolve. 
Overall, FOGSense achieved a 22.2\% improvement in F1-score and a 74.53\% reduction in false positive rate compared to state-of-the-art methods, along with enhanced sensitivity for FOG episode detection, empowering preventive care and long-term symptom management as Parkinson’s progresses. Code is available: \href{https://github.com/shovito66/FOGSense}{\textcolor{blue}{https://github.com/shovito66/FOGSense}}.

\end{abstract}

\maketitle
\vspace{-1mm}
\section{Introduction}
Freezing of Gait (FOG) is one of the most incapacitating symptoms of Parkinson's disease (PD), a neurodegenerative condition that impairs coordination and mobility. The transient inability to start or continue walking, a hallmark of FOG, significantly impairs a patient's mobility and quality of life, increasing the risk of falls, accidents, and major health concerns~\cite{brederecke2023freezing, lichter2021freezing}. Therefore, timely detection of FOG and effective interventions can save a PD patient from severe injuries or even death. Timely treatments, including external cues or therapy changes, can help reduce falls, increase mobility, and improve the overall quality of life when FOG is detected early~\cite{ahlrichs2016detecting}. Additionally, real-time detection and ongoing monitoring could keep a doctor well-informed and help modify treatment regimens, which can lessen the frequency and intensity of FOG episodes~\cite{borzi2021prediction}.

The integration of wearables with machine learning (ML) represents a major advancement in the continuous monitoring and management of PD symptoms. This convergence enables real-time motion pattern recognition, predictive modeling of FOG episodes, and personalized therapeutic insights.  However, the effectiveness of ML-driven FOG detection depends heavily on sensor placement, reliability, and adaptability, as capturing nuanced gait dynamics remains a challenge~\cite{tao2012gait}. While CyclePro~\cite{8616844} demonstrates the utility of wearable sensors for continuous gait cycle detection, especially in noisy environments, it also highlights the necessity for platform-independent solutions that do not require predefined parameters, which could greatly benefit FOG detection systems. SmartSock~\cite{fallahzadeh2016smartsock} offers a robust, wearable platform for continuous monitoring, using machine learning and signal processing techniques to provide real-time, context-aware assessments of ankle edema.

Multi-sensor systems, while providing richer data, introduce critical reliability concerns. Wearable FOG detection solutions that rely on multiple sensors at FoG-dominant locations (e.g., ankles~\cite{koltermann2024gait} or shoes~\cite{park2024detection}) are susceptible to sensor failures, misalignment, and increased system complexity. Sensor failures can decrease detection accuracy by up to 65\%~\cite{mamun2022designing}, especially in systems that rely on multiple channels, where each dependency increases the risk of system failure in real-world deployments.

Despite the advantages of wearables, FOG detection remains challenging due to the complex, non-linear nature of gait disturbances~\cite{liu2024adaptive}. Conventional time series frameworks struggle to capture the complex spatiotemporal dynamics and variability in gait movement, limiting their effectiveness in processing the multivariate data streams required for real-time FOG detection and intervention~\cite{pardoel2019wearable,Soumma_Mamun_Ghasemzadeh_2025}. 
To address this, we leverage Gramian Angular Field (GAF) transformations, a signal processing technique that converts temporal sequences into 2D representations and improves FOG detection~\cite{elmir2023ecg}. It outperforms wavelet and recurrence plot analysis by preserving local and global patterns while maintaining order~\cite{wang2015imaging}, boosting FOG detection accuracy~\cite{le2023gaformer}. 

To further investigate and enhance the efficacy of GAF and Convolutional Neural Network (CNN) for FOG detection, we designed FOGSense, a novel framework for FOG episode detection using only one axis of a single accelerometer sensor with a low false positive rate. This minimalist approach reduces hardware dependencies and enhances reliability while maintaining strong detection capabilities. By applying GAF transformations with a CNN, FOGSense enables both window-level and episode-level FOG detection. It effectively captures essential gait signatures within brief time windows, allowing for real-time use on resource-limited devices with low computational demand. Additionally, its single-axis design and dynamic weight transfer method improve precision and resilience against sensor failures.

Subsequently, we take another step to improve FOGSense by integrating federated learning. Modern smartphones can train models on-device but often strain battery and CPU~\cite{ignatov2019ai,9928465}. 
FOGSense overcomes this by employing a distributed learning framework tailored for wearable-smartphone ecosystems~\cite{wang2020convergence,khan2020federated}. This setup transfers sensor data during off-peak hours to capture FOG patterns without interrupting regular use~\cite{mazilu2015prediction}. A central server then uses TensorFlow to train models, converts them to TensorFlow Lite for deployment, and adapts these models to both individual and population-wide FOG variations~\cite{luo2020hfl,li2020federated,rieke2020future}. To achieve this, we explore the research question: \textbf{How can we reduce false positive (FP) intervention rates while maintaining high FOG detection accuracy, ensuring the model adapts to different patients over time without compromising privacy?}

In summary, our contributions are (i) a real-world deployable, privacy-preserving FOG detection system using a single lower-back accelerometer, reducing multi-sensor failure risks, (ii) Integration of GAF transformations and federated learning, enhancing gait pattern recognition with lower FP rate (iii) the development and evaluation of the FOGSense system for window-level, episode-level, and federated FOG detection, (iv) performance comparison across channels and with existing methods, and (v) a dynamic weight transfer framework to address channel failure.
\section{Related Work}
Recent advancements in FOG detection address the challenges of traditional sensor-based ML methods, with key directions being (1) raw sensor-based deep learning, (2) self-supervised learning (SSL), and (3) federated learning.

Raw sensor-based deep learning minimizes manual feature engineering by applying models like 1D-CNNs directly to accelerometer data~\cite{zhao2018hybrid}. 
These methods reduce preprocessing overhead but require large labeled datasets and often fail to capture the full complexity of sensor signals. Moreover, they struggle with generalizability, particularly when faced with variations in patient populations or sensor placements~\cite{sama2012dyskinesia}, limiting their real-world applicability. Recent works like DGAD ~\cite{naghavi2021towards}, FoG-Finder~\cite{koltermann2023fog} and Gait-Guard~\cite{koltermann2024gait} use two ankle-worn IMUs and CNN-based models for real-time FoG detection. DGAD applies a Butterworth bandpass filter (0.5-8Hz) to IMU data before using a single-headed CNN for classification. While some of these systems achieve strong detection performance, they rely on more than one sensor, increasing hardware complexity and failure risks, and are tested in a controlled setting.

SSL techniques aim to overcome labeled data constraints by learning meaningful representations from raw data~\cite{soumma2024self,xia2024prediction}. These methods extract essential features without extensive annotation but face challenges in adapting to diverse sample distributions. Variations in environmental conditions, sensor noise, and individual gait patterns can significantly affect performance~\cite{lonini2018wearable}, highlighting the need for robust adaptation mechanisms.

Federated learning enables collaborative model training across devices while preserving data privacy~\cite{li2020federated,chen2020fedhealth}. This distributed approach is particularly suitable for healthcare but presents challenges in resource-constrained wearables, such as balancing computational efficiency, power consumption, and model performance.

Building on these foundations, our work integrates GAF transformation with multichannel CNNs for enhanced feature extraction, surpassing traditional raw sensor-based approaches. Additionally, our federated learning implementation improves model adaptability while preserving privacy and addressing efficiency challenges, making it suitable for resource-constrained real-world applications.

\section{FOGSense System Design}
\subsection{Federated Learning on Edge Devices}
\begin{figure}[!h]
     \vspace{-2mm}
    \centering
    \includegraphics[width=1\linewidth]{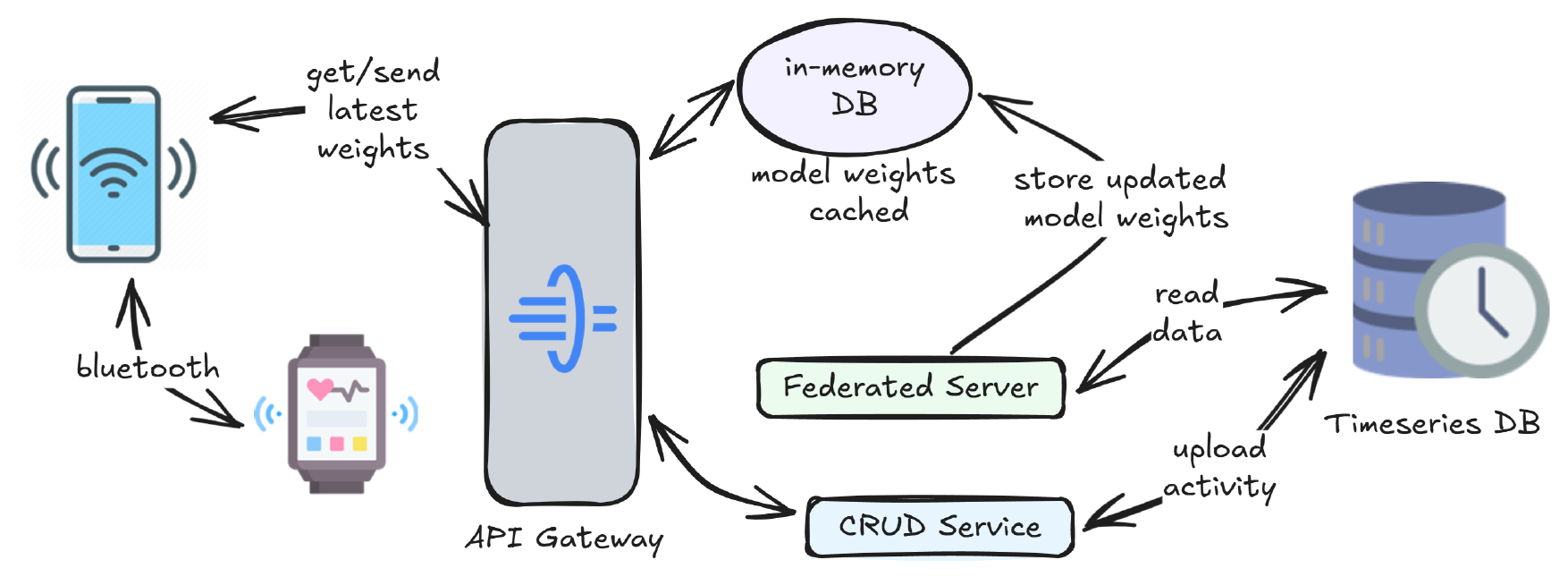}
    \caption{Proposed distributed federated system design.}
    \label{fig:federated_system_design}
\end{figure}

\begin{figure*}
    \centering
    \includegraphics[width=0.94\linewidth, trim={0 150 0 50}, clip]{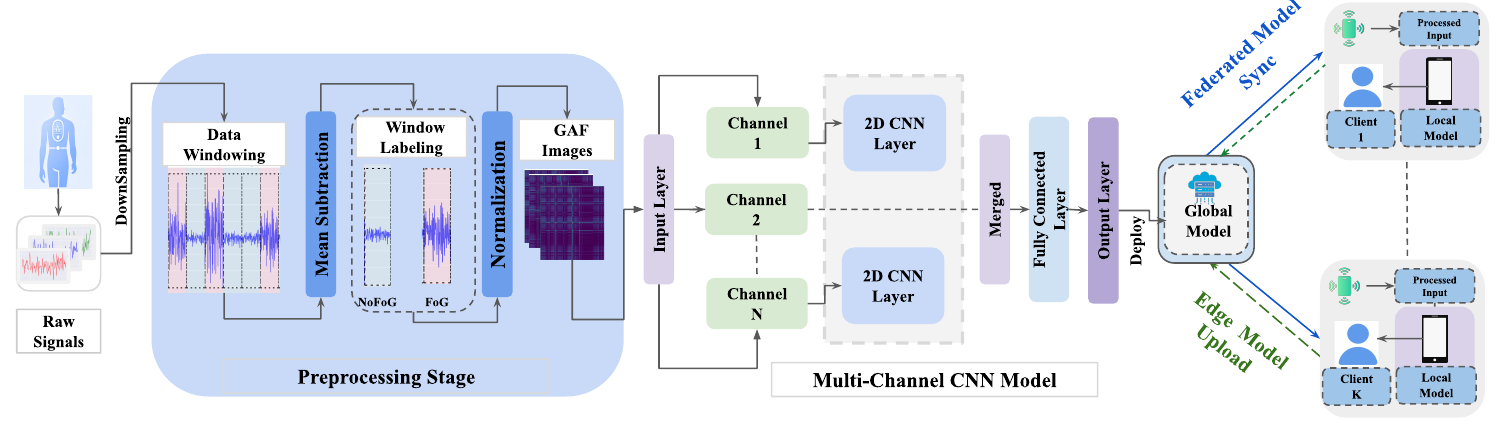}
    \vspace{-2mm}
    \caption{System architecture illustrating the federated learning workflow: local models are trained and uploaded by devices, then aggregated into a global model, which is distributed back to each device.}
    \label{fig:fogsense_Design}
    \vspace{-5mm}
\end{figure*}

Developing a reliable FOG detection model is challenging due to the complexity of PD and symptom variability across patients. Our hierarchical system architecture, illustrated in Fig.~\ref{fig:federated_system_design}, addresses these challenges while accommodating the constraints of wearable devices. Wearables continuously collect acceleration data, which is transferred to paired smartphones at designated intervals. A central federated server aggregates this data during off-peak hours, applies GAF transformations, and executes model training as shown in Fig.~\ref{fig:fogsense_Design}. The processed data are stored in a time-series database for comprehensive health monitoring and temporal analysis. Updated model weights are saved in an in-memory database for efficient synchronization with smartphones, which then update wearables. This design ensures efficient data handling, robust model distribution, and reliable FOG monitoring. 

We trained a multi-channel CNN model across five federated clients, each simulating different data distributions. Clients were assigned different combinations of gait data, allowing the system to learn personalized gait variations without centralizing sensitive patient data. We used a weighted averaging strategy (FedAvg) to aggregate client updates~\cite{beaussart2021waffle}, ensuring a balanced contribution to the global model while preserving model stability. Each client trained on local epochs before synchronizing with the central server, which converted the models to TensorFlow Lite for efficient deployment on resource-constrained devices. This approach enables continuous, personalized adaptation of FOG detection while maintaining low communication overhead and strong privacy guarantees.

\subsection{Dataset}
We used the publicly available PD dataset ~\cite{tlvmc-parkinsons-freezing-gait-prediction} from the TLVMC FOG prediction competition. The dataset has 62 subjects diagnosed with Parkinson’s disease, specifically focusing on FOG episodes. The analysis concentrated on the `tdcsfog' dataset. Data were collected via a lower-back triaxial accelerometer at 128 Hz, capturing vertical (AccV), mediolateral (AccML), and anteroposterior (AccAP) movements, critical for identifying directional mobility loss during FOG events. Rich temporal annotations, including start hesitations, turns, and walking, enabled precise FOG analysis. Using the ``Differential Hopping Windowing Technique'' (DHWT)~\cite{soumma2024self}, the FOG window counts rose from 1,096,265 to 1,994,705, yielding an 82\% increase.
\begin{figure}[H]
    \vspace{-5pt}
    \centering
    \includegraphics[width=0.5\textwidth,trim={0 60 0 20}, clip]{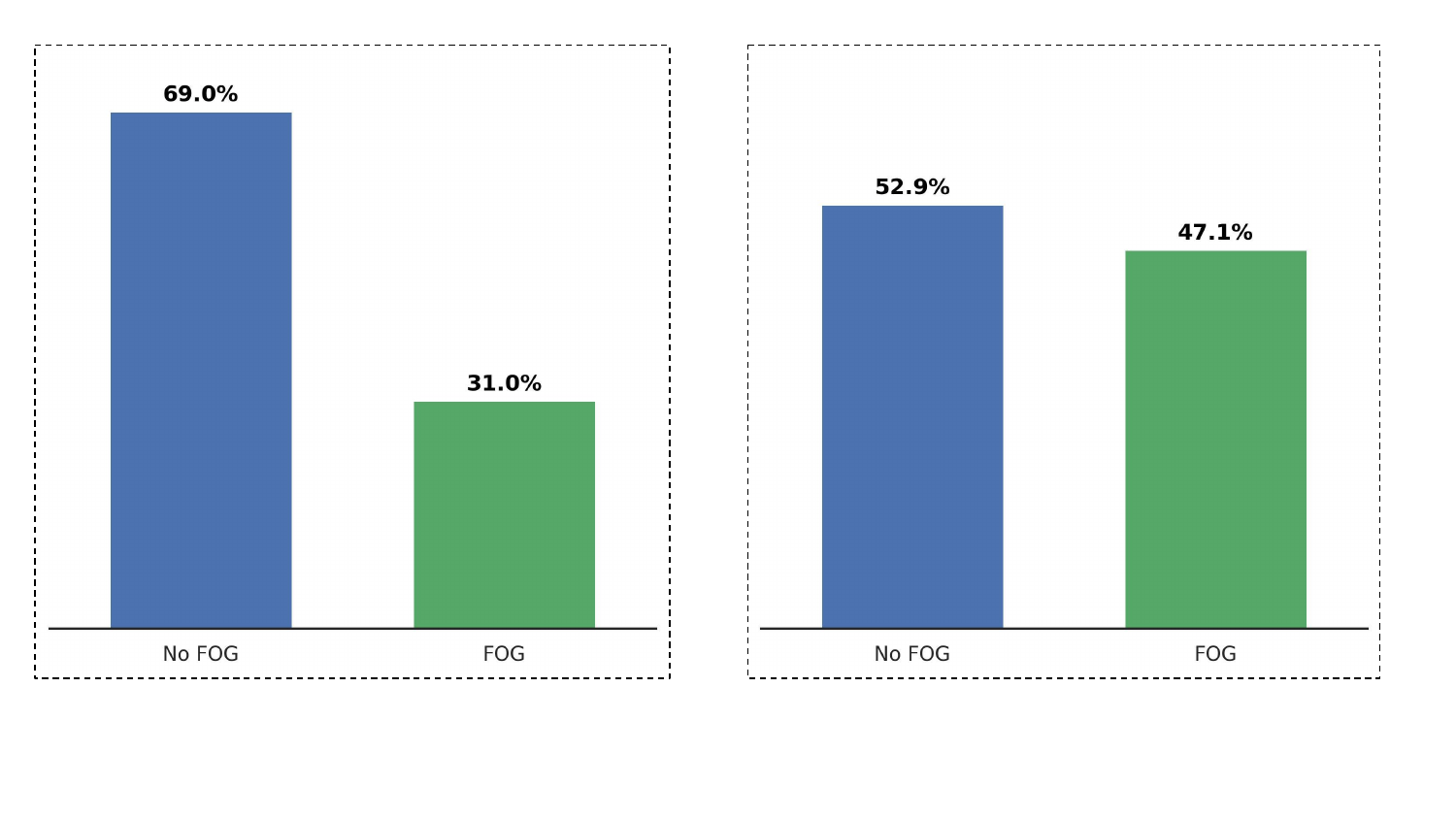}
    \begin{center}
       \textbf{(a) Before DHWT} \hspace{0.8cm} \textbf{(b) After DHWT}  
    \end{center}
    
    \vspace{-4pt}
    \caption{Class distribution normal and FOG events before (a) and after (b) applying (DHWT) to the training set.}
    \label{fig:fog_event_distribution}
    \vspace{-10pt}
\end{figure}

\subsection{Dataset Stratification}
We used subject-based stratified partitioning to prevent data leakage and ensure robust evaluation.
The dataset was split into training (70\%), validation (10\%), and testing (20\%) sets using randomly shuffled subject identifiers and it preserved FOG/no-FOG proportions and temporal integrity~\cite{hammerla2016deep}. This procedure is repeated for each group within the dataset, conducted three times, and the results are averaged to ensure reliability and robustness in our findings.

\subsection{Pre-processing}
The dataset, originally collected at 128Hz, was downsampled to 64Hz to reduce computational complexity while preserving signal characteristics~\cite{soumma2024self}. The `tdcsfog' data's three event types were consolidated into a single column, labeling FOG as 1 and no-FOG as 0.

A critical preprocessing step involved defining time windows for analysis, which we set at 4-second intervals to effectively capture essential gait features for FOG detection~\cite{naghavi2019prediction}. To address class imbalance effectively, we used DHWT~\cite{soumma2024self} on our training set. This technique applied varying overlap sizes: no overlap (0\%) for no-FOG windows and an increased 50\% overlap for FOG windows. By using this method, we enhanced the representation of FOG events, mitigating class imbalance and enriching the model's training dataset without additional preprocessing. Fig.~\ref{fig:fog_event_distribution} illustrates the results achieved with this approach, showing a more balanced distribution of FOG and no-FOG instances. We filtered out impure windows where no-FOG events (class 0) predominated, designating windows where FOG events (class 1) were in the majority as pure FOG windows. For improved data consistency and distribution, we subtracted the mean value from each window ($X_{i,j}^{\prime} = X_{i,j} - \mu_{X_i}$), ensuring normalization across samples.

Our methodology employs Conv2D CNN for FOG episode detection. To leverage CNNs' proven capabilities in image processing, we transformed our temporal acceleration data into image-based representations using the GAF method. This transformation encodes temporal dependencies into a higher-dimensional space, allowing the CNN architecture to potentially capture complex patterns that might be less apparent in the original time series format.

\subsection{Gramian Angular Fields}
The GAF transformation effectively converts 1D time series data into 2D images suitable for CNN-based analysis while preserving temporal and dynamic structures~\cite{yokkampon2022autoencoder}. The transformation process consists of two key steps: normalization and polar coordinate mapping. First, the time series data is normalized to the range $[-1,1]$ through the following transformations:
\begin{align}
x'_i &= \frac{x_i - \min(X)}{\max(X) - \min(X)} &
\theta_i &= \cos^{-1}(x'_i) \label{eq:angle} \\[6pt]
x''_i &= 2x'_i - 1 &
r_i &= \frac{t_i}{N} 
\label{eq:radius}
\end{align}

where $t_i$ represents the time stamp and $N$ denotes the total number of points.
\begin{figure}[H]
\vspace{-8pt}
    \centering
        \includegraphics[width=1\linewidth, trim={20 170 10 60}, clip]{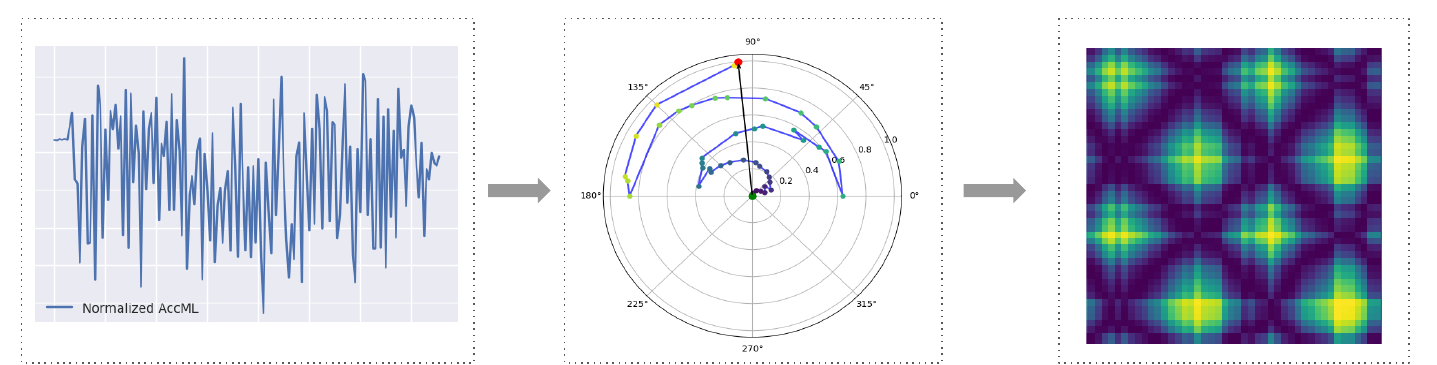}
    \begin{center}
        \vspace{-2mm}
        \hspace{0.05cm}(a) Time series \hspace{0.65cm} (b) Polar coordinates  \hspace{0.2cm} (c) GAF image
    \end{center}
    \vspace{-5pt}
\caption{GAF transformation of accelerometer signals.}
    \label{fig:Transform_image}
    \vspace{-8pt}
\end{figure}

The next stage in the GAF process involves creating the Gram matrix, which captures temporal correlations between different time points using trigonometric functions. Specifically, for the Gramian Angular Summation Field (GASF), we compute:
\begin{equation}
    \text{GASF}(i,j) = \cos(\theta_i + \theta_j)
    \label{eq:gasf}
\end{equation}

For a time series of $m$ data points, the resulting GASF matrix takes the form:
\begin{equation}
    \text{GAF} = \begin{bmatrix}
    \cos(\theta_1 + \theta_1) & \cdots & \cos(\theta_1 + \theta_m) \\
    \vdots & \ddots & \vdots \\
    \cos(\theta_m + \theta_1) & \cdots & \cos(\theta_m + \theta_m)
    \end{bmatrix}
    \label{eq:gaf_matrix}
\end{equation}

\begin{figure}[!h]
   \vspace{-10pt}
   \centering
   \includegraphics[width=0.48\textwidth,trim={0 0
 10 10}, clip]{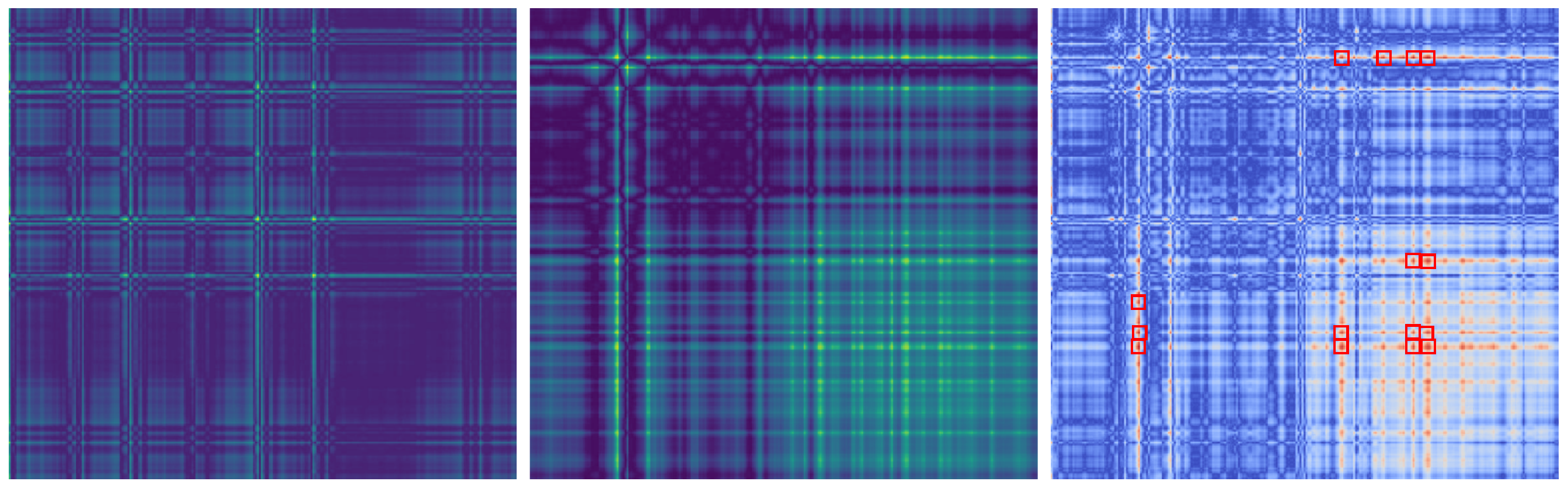}
    \begin{center}
        \small
        \hspace{0.1cm}(a) No-FOG image \hspace{0.7cm} (b) FOG image \hspace{0.71cm} (c) Differences
    \end{center}
    \vspace{-8pt}
   \caption{GAF-transformed images for (a) No-FOG, (b) FOG conditions, and (c) differences, with significantly different regions highlighted with red boxes.}
   \vspace{-5pt}
\label{fig:Diffrence_image}
\end{figure}
As shown in Fig.~\ref{fig:Diffrence_image}, the GAF transformation creates distinct visual patterns for normal (Fig.~\ref{fig:Diffrence_image}a) and FOG (Fig.~\ref{fig:Diffrence_image}b) conditions, enabling image-based deep learning applications. The annotated difference, based on pixel-wise absolute differences between GAF-transformed images, highlights areas with significant pattern variations, capturing subtle gait changes. The top 15 regions with the highest difference values are emphasized, showcasing where the FOG pattern deviates most from no-FOG. This approach effectively captures temporal patterns in sequential data, making it ideal for medical signal processing and gait analysis \cite{buz2020novel}.

\subsection{Experimental Setup}
\label{subsec:exp_setup}
FOGSense uses a multichannel 2D-CNN architecture designed to maximize performance through spatial relationship analysis across multiple sensor modalities. The model processes data from three distinct accelerometer axes (vertical, mediolateral, and anteroposterior) by transforming them into GAF images and processing them through dedicated CNN branches. This approach enables the system to leverage complementary information from different sensor perspectives, significantly enhancing its FOG detection capabilities. All experiments, including pre-processing, model training, and federated learning, were conducted using Python 3.8, TensorFlow 2.2, and one NVIDIA A100 GPU.

\begin{figure}[!h]
   \centering
   \includegraphics[width=\linewidth ,trim={60 100 80 130}, clip]{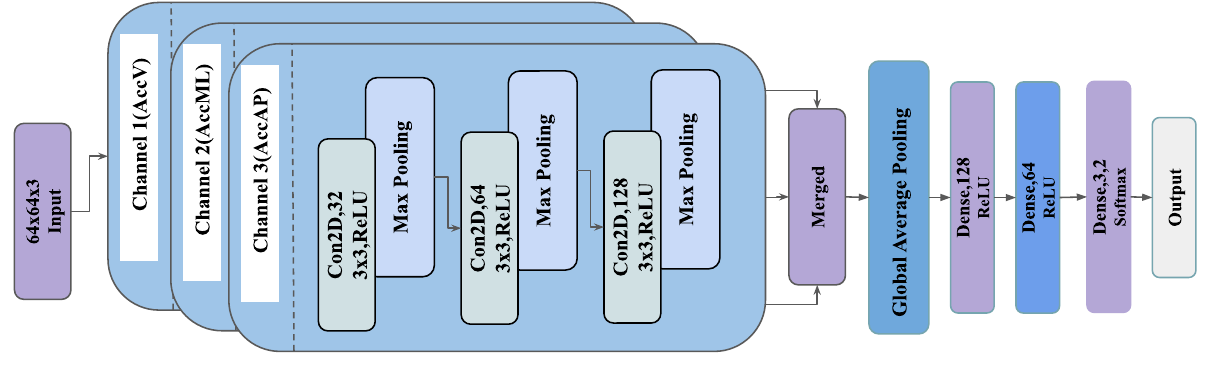}
\vspace{-4mm}
   \caption{The multichannel CNN architecture.}
   \label{fig:cnn_image}
   \vspace{-2mm}
\end{figure}

The proposed architecture (Fig.~\ref{fig:cnn_image}) consists of three parallel CNN branches, each processing GAF images from a specific accelerometer axis. Each branch processes input data with dimensions (64, 64, 3) through three consecutive convolutional blocks with increasing complexity - starting with 32 filters (20\% dropout), followed by 64 and 128 filters, incorporating L2 regularization ($\lambda=0.001$) and dropout rates of 20\% and 40\%.
\begin{table}[!h]
    \vspace{-2.5pt}
    \centering
    \caption{Dense layers after feature concatenation.}
    \vspace{-2mm}
    \begin{tabular}{lccc}
    \toprule
    \textbf{Layer} & \textbf{Units} & \textbf{Regularization} & \textbf{Dropout} \\ 
    \midrule
    First Dense  & 128 & L2 ($\lambda=0.001$) & 40\% \\ 
    Second Dense & 64 & None & 20\% \\ 
    Output Dense & 2 & None & None \\ 
    \bottomrule
    \end{tabular}
    \vspace{-2mm}
    \label{tab:dense_layers}
\end{table}
Batch normalization and max pooling enhance feature extraction while reducing spatial dimensions.

A global average pooling layer further compresses parameters before features from all branches are concatenated. The final classifier consists of two dense layers (128 and 64 neurons with ReLU activation, dropout (40\%, 20\%), and L2 regularization) followed by a softmax layer for FOG classification. Adam optimizer and categorical cross-entropy loss are used to improve generalization. The backbone architecture remained unchanged during channel configuration analysis to ensure consistent performance evaluation.

Finally, we also proposed a dynamic weight transfer mechanism to handle the failure of any sensor channel. As the architecture is the same for each channel, when a channel fails, the system can load the weights for a different functional channel and predict with that channel. For example, if the channel AccV fails at any time and the channel AccAP is functional, the FOGSense system will load the weights trained with AccAP and make the prediction. This is shown in Algorithm~\ref{alg:dynamic_weight_transfer}. 

\subsection{Performance Measures}
FOGSense was evaluated using standard binary classification metrics. Based on the confusion matrix elements—true positives (TP), false positives (FP), false negatives (FN), and true negatives (TN)—we computed the following metrics: sensitivity, precision, accuracy, F1-score, and false positive rate (FPR). FPR is crucial in FOG detection as it quantifies incorrect intervention triggers, ensuring the system minimizes unnecessary alerts; otherwise, treatment efficacy will decrease.
\vspace{-5pt}

\begin{algorithm}[h]
\caption{Inference with Dynamic Weight Transfer}
\begin{algorithmic}[1]
    \State \textbf{Training and testing before deployment:}
    \State Train $N$ models of identical architecture separately for $N$ channels and evaluate them on the test dataset.
    \State Sort the $N$ models and channels in descending order of their test F-1 scores as:
    \Statex \hspace{2em} $\text{Models} = (\text{Model}_1, \text{Model}_2, \dots, \text{Model}_N)$
    \Statex \hspace{2em} $\text{Channels} = (\text{Channel}_1, \text{Channel}_2, \dots, \text{Channel}_N)$
    \State Save the weights of all models.
    
    \State \textbf{Inference after deployment:}
    \State Find the smallest $i$ such that $\text{Channel}_i$ is functional in new data.
    \State Load weights of $\text{Model}_i$.
    \State Predict using the $i$-th channel's readings in the new data.
\end{algorithmic}
\label{alg:dynamic_weight_transfer}
\end{algorithm}

    
\vspace{-5mm}
\section{Results}
To ensure the integrity of GAF-transformed signals, we did not apply conventional data augmentation techniques. We evaluated FOGSense for both window-level and episode-level detection, achieving consistently high performance. Window-level detection refers to independently classifying each window as FOG or non-FOG and episode-level detection merges contiguous windows to form a single FOG event. To improve clinical applicability, we implemented federated learning, allowing continuous model adaptation while preserving privacy.
Table~\ref{tab:performance_summary} shows that FOGSense achieves high accuracy across all detection levels, with F1-scores above 90\% and a low false positive rate (3.2-3.8\%). Episode-level detection performs best, minimizing false alarms.
\begin{table}[!h]
\centering
\caption{FOGSense's performance for three different detection types using all three channels. DFW: Detected FOG Windows, DFE: Detected FOG Episodes, FPR: False Positive Rate.}
\setlength{\tabcolsep}{3.9pt}  
\renewcommand{\arraystretch}{0.8}  
\resizebox{\linewidth}{!}{%
\begin{tabular}{lcccc}
\toprule
\textbf{Detection Type} & \textbf{Configuration} & \textbf{Accuracy (\%)} & \textbf{F1 (\%)}  &\textbf{FPR (\%)}\\
\midrule
Window-level (DFW) & All Channels & 87.23 & 90.31 &3.8\\
\midrule
Episode-level (DFE) & All Channels & 87.23 & 90.86 &3.2\\
\midrule
Federated Learning & All Channels & 86.98 & 90.47 &3.5\\
\bottomrule
\end{tabular}%
}
\label{tab:performance_summary}
\vspace{-2mm}
\end{table}


\subsection{Effect of Channel Selection}
We conducted a systematic evaluation of input channel configurations to optimize our FOG detection framework. Initially utilizing all three accelerometer axes (anterior-posterior, mediolateral, and vertical), we investigated the impact of channel reduction on model performance. The analysis focused on quantifying the trade-off between computational efficiency and FOG detection accuracy across different channel combinations to determine the optimal configuration. Here, each model was trained for 60 epochs.

\vspace{-1mm}
\begin{table}[!h]
    \centering
    \caption{Impact of channel selection}
    \vspace{-1mm}
    \setlength{\tabcolsep}{2.5pt}
    \scalebox{0.9}{
\begin{tabular}{cccccccc}
    \toprule
     \multirow{2}{*}{\textbf{Channels}} & \multirow{2}{*}{\makecell{\textbf{Input}\\\textbf{Combination}}} &\multicolumn{2}{c}{\textbf{Duration (s/epoch)}} &\multirow{2}{*}{\textbf{ACC}} &\multirow{2}{*}{\textbf{F1}} &\multirow{2}{*}{\textbf{PRE}} &\multirow{2}{*}{\textbf{SEN}}\\ 
     \cmidrule(l){3-4}
     &&Train &Test \\
    \midrule
         3 &AccAP-AccML-AccV & 63.0 & 12.9 & 87.2 & 90.3 & 90.3 & 91.0 \\
         \hline
         \multirow{3}{*}{\centering 2} & 
    AccAP-AccML & 32.1 & 15.4 & 93.2 & 87.7 & 93.2 & 93.2 \\
    & AccML-AccV & 31.8 & 19.0 & 95.3 & 90.8 & 95.3 & 95.3 \\
    & AccAP-AccV & 31.7 & 13.3 & 94.5 & 90.5 & 94.5 & 94.5 \\
    \midrule
    \multirow{3}{*}{\centering 1} &
    AccAP & 11.3 & 11.0 & 93.2 & 93.0 & 93.2 & 93.2 \\
    & AccML & 17.7 & 19.0 & 89.2 & 83.0 & 89.2 & 89.2 \\
    & \textbf{AccV} & 11.3 & 11.0 & \textbf{96.3} & \textbf{96.3} & \textbf{95.8} & \textbf{96.3} \\
    \bottomrule
\end{tabular}}
    \label{tab:something-else}
    \vspace{-2mm}
\end{table}

Table~\ref{tab:something-else} reveals several key insights regarding channel configuration impact on model performance. The baseline three-channel configuration (AccAP-AccML-AccV) achieved 87.23\% accuracy and 90.31\% F1-score, requiring 63 seconds per training epoch. However, reducing to two channels generally improved performance while halving the training time to approximately 32 seconds per epoch. Among dual-channel configurations, AccML-AccV demonstrated superior performance with 95.30\% accuracy and 90.75\% F1-score.

Notably, single-channel configurations further reduced training time to approximately 11 seconds per epoch while maintaining or improving performance. The AccV channel alone achieved the highest overall accuracy of 96.25\% and F1-score of 96.27\%, surpassing both three-channel and two-channel configurations. In contrast, AccML showed a relatively lower performance (89.24\% accuracy), while AccAP maintained a robust performance (93.24\% accuracy).
\begin{figure}[!h]
   \centering
   \includegraphics[width=1\linewidth, trim={85 0 100 0}, clip]{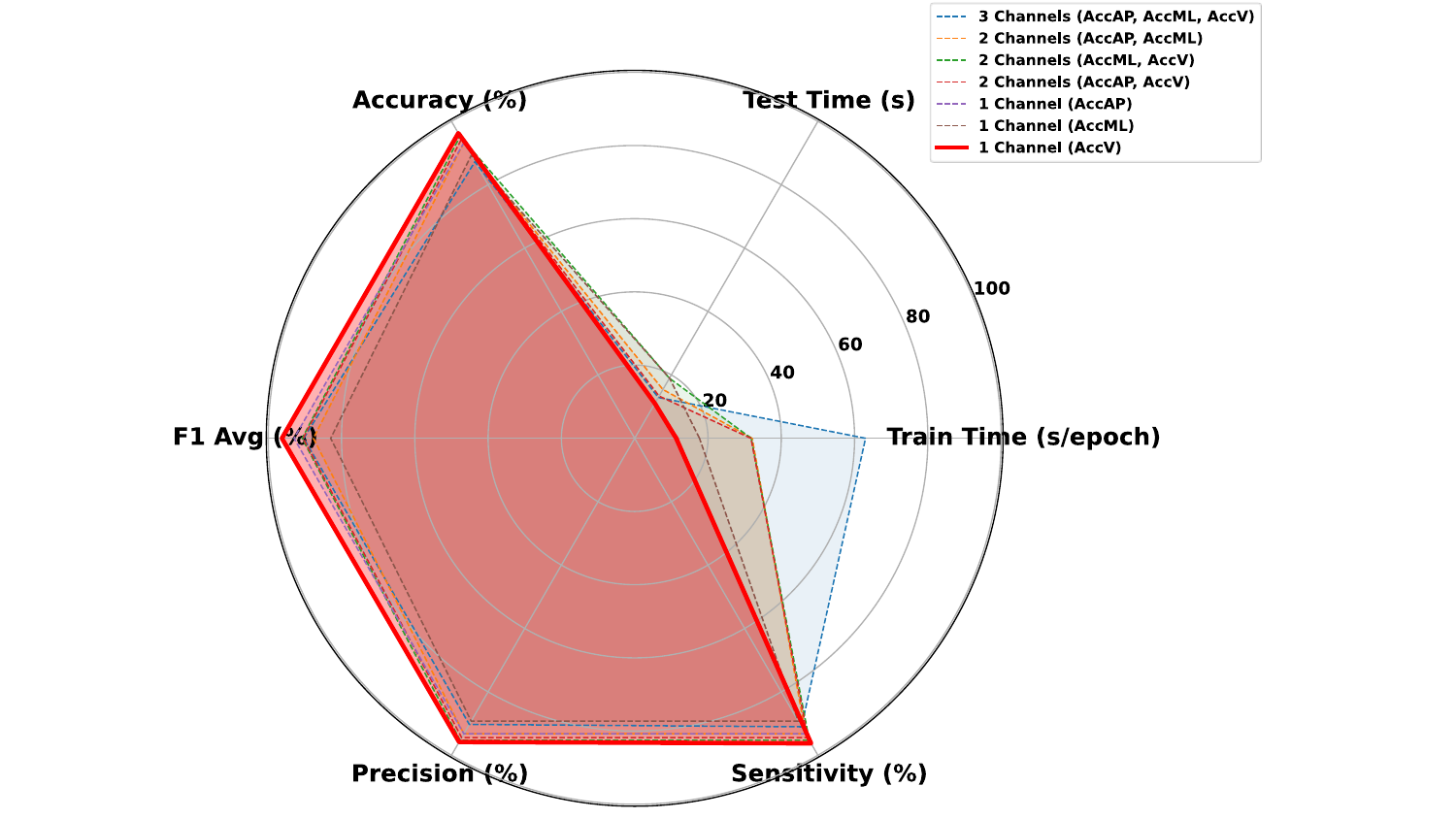}
   \caption{Comparison of channel configurations in the FOGSense system shows that the vertical axis (AccV) obtains a better F1 score than other channel choices.}
   \label{fig:channel_comparison_radar}
   \vspace{-9pt}
\end{figure}
As illustrated in Fig.~\ref{fig:channel_comparison_radar}, the single AccV channel configuration (highlighted in red) emerges as the optimal choice, offering superior performance metrics while minimizing computational overhead. This finding suggests that AccV alone captures sufficient information for effective FOG detection, making it particularly suitable for resource-constrained applications. Additionally, this configuration is resilient to missing values, as it continues to perform effectively even when data from other axes is unavailable, achieving better results using only the AccV channel.

\subsection{Comparison with State-of-the-Art}
We present a comparative analysis of FOGSense and the other state-of-the-art (SOTA) methods in Table~\ref{tab:performance-compare}. Our FOGSense method achieves an F1 score of 0.96 on the `tdcsfog' dataset, whereas the closest competitor, LIFT-PD achieved an F1 score of 0.79. Moreover, unlike the other methods, the FOGSense maintains a score of at least 0.90 in all four metrics, with an average 74.5\% reduction in FPR. Additionally, FOGSense's use of the vertical axis (AccV) channel selection not only enhances detection accuracy but also provides a more efficient approach to gait monitoring. These results underscore FOGSense's potential as a powerful and dependable tool for real-time gait analysis. 

\begin{table}[h]
    \vspace{-1pt}
    \caption{FOGSense vs. state-of-the-art on \textit{tdcsfog}. The best values are highlighted in bold, while the second-best values are underlined.}
    \centering
    \vspace{-2pt}
    \setlength{\tabcolsep}{2.5pt} 
    \renewcommand{\arraystretch}{0.9} 
    \scalebox{1}{
    \begin{tabular}{c|*{3}{c} cc}
    \toprule
         \textbf{Study} & \textbf{DFE}   & \textbf{SEN (DFW)} & \textbf{PRE} & \textbf{F1} &\textbf{FPR}\\
         \midrule
         One Class Classifier~\cite{naghavi2021towards}         &90.8\% &71.6\%       &\underline{0.86 }  &0.77  &12.4\%\\
         Semi-Supervised Model~\cite{mikos2017real} & -- & 72.3\% &--&--  &11.8\% \\
         Multi-head CNN~\cite{10.1016/j.artmed.2022.102459}& 97.3\%&\underline{94.6}\%      &0.56   &0.68   &20.5\% \\
         LIFT-PD~\cite{soumma2024self}        &88\%   &{84\%}         &{0.74}   &\underline{0.79} &\underline{9.6}\% \\
         \midrule
         \textbf{FOGSense}          &--   &\textbf{96.3\%}         &\textbf{0.958}   &\textbf{0.963} &\textbf{3.2}\%\\
         \bottomrule
    \end{tabular}}
    \vspace{-2mm}
    \label{tab:performance-compare}
\end{table}
\section{Discussion}
In this paper, we introduced a novel approach to detect FOG by integrating GAF-transformed accelerometer data with a multichannel CNN architecture and a decentralized federated learning pipeline. Our experiments demonstrated not only high detection accuracy with a low false positive intervention rate but also highlighted the practicality of our system for in-home real-time monitoring. A key finding is that a single-channel configuration using only the vertical acceleration component (AccV) achieved impressive performance—96.25\% accuracy and 96.27\% F1-score—while significantly reducing computational demands compared to multi-channel configurations. This result challenges the traditional view that multiple sensor channels are essential for accurate FOG detection while ensuring a balance between high accuracy and minimal false positives, preventing unnecessary interventions that could reduce treatment efficacy over time.

To address continuous monitoring constraints, we developed a distributed framework that synchronizes data between wearable devices and smartphones during off-peak hours, balancing the demands of monitoring, battery life, and computational efficiency. Although the GAF transformation was effective in our approach, exploring alternative time series representations such as Recurrence Plots and Markov Transition Fields may yield further insights. Additionally, while our distributed framework shows promise in simulations, real-world validation on multiple edge devices is a critical next step. Investigating different weight calculation methods beyond weighted averaging could enhance the learning dynamics in federated learning. Testing the model across diverse patient populations and datasets would further strengthen its generalizability, and studying patient data mixing strategies may provide insights into managing individual gait variations effectively.

\section{Conclusion}
Accurate detection of freezing of gait (FOG) can significantly enhance early intervention strategies, improve mobility support, and ultimately contribute to a better quality of life for individuals with Parkinson’s disease. Our FOGSense system uses Gramian Angular Field (GAF) imaging, CNN, and federated learning to ensure accurate and resource-efficient FOG detection in an uncontrolled environment. Moreover, FOGSense maintains reliability even with sensor failure, as it needs only one channel to make an accurate prediction and it can delegate the prediction task to a different single-channel model by dynamic weight transfer in case of a channel failure.

Our work establishes a foundation for efficient, practical, and reliable FOG detection systems. We provide new insights into lightweight, single-channel solutions that outperform multi-channel approaches, achieving a 22.2\% improvement in F1-score and a 74.5\% reduction in false positive rate (FPR), minimizing unnecessary interventions that could diminish their therapeutic impact over time. By reducing false positives, FOGSense ensures that patients receive interventions only when needed, preventing desensitization to stimuli and maintaining long-term clinical effectiveness. We also identified future research directions that could advance the field, from alternative data representations to federated learning enhancements and real-world validations. This study highlights the potential of FOGSense and opens avenues for refining real-time FOG detection systems to meet clinical and practical demands more effectively.
\vspace{-1.5mm}

\bibliographystyle{ieeetr}
\bibliography{sample.bib}

\begin{thebibliography}{10}

\bibitem{brederecke2023freezing}
J.~Brederecke, ``Freezing of gait prediction from accelerometer data using a simple 1d-convolutional neural network--8th place solution for kaggle's parkinson's freezing of gait prediction competition,'' {\em arXiv preprint arXiv:2307.03475}, 2023.

\bibitem{lichter2021freezing}
D.~G. Lichter, R.~H.~B. Benedict, and L.~A. Hershey, ``Freezing of gait in parkinson’s disease: risk factors, their interactions, and associated nonmotor symptoms,'' {\em Parkinson’s Disease}, vol.~2021, no.~1, p.~8857204, 2021.

\bibitem{ahlrichs2016detecting}
C.~Ahlrichs, A.~Sam{\`a}, M.~Lawo, J.~Cabestany, D.~Rodr{\'\i}guez-Mart{\'\i}n, C.~P{\'e}rez-L{\'o}pez, D.~Sweeney, L.~R. Quinlan, G.~{\`O}. Laighin, T.~Counihan, {\em et~al.}, ``Detecting freezing of gait with a tri-axial accelerometer in parkinson’s disease patients,'' {\em Medical \& biological engineering \& computing}, vol.~54, pp.~223--233, 2016.

\bibitem{borzi2021prediction}
L.~Borz{\`\i}, I.~Mazzetta, A.~Zampogna, A.~Suppa, G.~Olmo, and F.~Irrera, ``Prediction of freezing of gait in parkinson’s disease using wearables and machine learning,'' {\em Sensors}, vol.~21, no.~2, p.~614, 2021.

\bibitem{tao2012gait}
W.~Tao, T.~Liu, R.~Zheng, and H.~Feng, ``Gait analysis using wearable sensors,'' {\em Sensors}, vol.~12, no.~2, pp.~2255--2283, 2012.

\bibitem{8616844}
Y.~Ma, Z.~Esna~Ashari, M.~Pedram, N.~Amini, D.~Tarquinio, K.~Nouri-Mahdavi, M.~Pourhomayoun, R.~D. Catena, and H.~Ghasemzadeh, ``Cyclepro: A robust framework for domain-agnostic gait cycle detection,'' {\em IEEE Sensors Journal}, vol.~19, no.~10, pp.~3751--3762, 2019.

\bibitem{fallahzadeh2016smartsock}
R.~Fallahzadeh, M.~Pedram, and H.~Ghasemzadeh, ``Smartsock: A wearable platform for context-aware assessment of ankle edema,'' in {\em 2016 38th Annual International Conference of the IEEE Engineering in Medicine and Biology Society (EMBC)}, pp.~6302--6306, IEEE, 2016.

\bibitem{koltermann2024gait}
K.~Koltermann, J.~Clapham, G.~Blackwell, W.~Jung, E.~N. Burnet, Y.~Gao, H.~Shao, L.~Cloud, I.~Pretzer-Aboff, and G.~Zhou, ``Gait-guard: Turn-aware freezing of gait detection for non-intrusive intervention systems,'' in {\em 2024 IEEE/ACM Conference on Connected Health: Applications, Systems and Engineering Technologies (CHASE)}, pp.~61--72, IEEE, 2024.

\bibitem{park2024detection}
J.-M. Park, C.-W. Moon, B.~C. Lee, E.~Oh, J.~Lee, W.-J. Jang, K.~H. Cho, and S.-H. Lee, ``Detection of freezing of gait in parkinson's disease from foot-pressure sensing insoles using a temporal convolutional neural network,'' {\em Frontiers in aging neuroscience}, vol.~16, p.~1437707, 2024.

\bibitem{mamun2022designing}
A.~Mamun, S.~I. Mirzadeh, and H.~Ghasemzadeh, ``Designing deep neural networks robust to sensor failure in mobile health environments,'' in {\em 2022 44th Annual International Conference of the IEEE Engineering in Medicine \& Biology Society (EMBC)}, pp.~2442--2446, IEEE, 2022.

\bibitem{liu2024adaptive}
Y.~Liu, X.~Liu, Q.~Zhu, Y.~Chen, Y.~Yang, H.~Xie, Y.~Wang, and X.~Wang, ``Adaptive detection in real-time gait analysis through the dynamic gait event identifier,'' {\em Bioengineering}, vol.~11, no.~8, p.~806, 2024.

\bibitem{pardoel2019wearable}
S.~Pardoel, J.~Kofman, J.~Nantel, and E.~D. Lemaire, ``Wearable-sensor-based detection and prediction of freezing of gait in parkinson’s disease: a review,'' {\em Sensors}, vol.~19, no.~23, p.~5141, 2019.

\bibitem{Soumma_Mamun_Ghasemzadeh_2025}
S.~B. Soumma, A.~Mamun, and H.~Ghasemzadeh, ``Domain-informed label fusion surpasses llms in free-living activity classification (student abstract),'' {\em Proceedings of the AAAI Conference on Artificial Intelligence}, vol.~39, pp.~29495--29497, Apr. 2025.

\bibitem{elmir2023ecg}
Y.~Elmir, Y.~Himeur, and A.~Amira, ``Ecg classification using deep cnn and gramian angular field,'' in {\em 2023 IEEE Ninth International Conference on Big Data Computing Service and Applications (BigDataService)}, pp.~137--141, IEEE, 2023.

\bibitem{wang2015imaging}
Z.~Wang and T.~Oates, ``Imaging time-series to improve classification and imputation,'' {\em arXiv preprint arXiv:1506.00327}, 2015.

\bibitem{le2023gaformer}
T.-H. Le, T.-K. Nguyen, T.-K. Tran, T.-H. Tran, and C.~Pham, ``Gaformer: Wearable imu-based human activity recognition with gramian angular field and transformer,'' in {\em 2023 Asia Pacific Signal and Information Processing Association Annual Summit and Conference (APSIPA ASC)}, pp.~297--303, IEEE, 2023.

\bibitem{ignatov2019ai}
A.~Ignatov, R.~Timofte, W.~Chou, K.~Wang, M.~Wu, T.~Hartley, and L.~Van~Gool, ``Ai benchmark: Running deep neural networks on android smartphones,'' in {\em Proceedings of the European Conference on Computer Vision (ECCV) Workshops}, pp.~0--0, 2019.

\bibitem{9928465}
S.~V.~P. Venkata, S.~Sabat, C.~A. Deshpande, A.~Arefeen, D.~Peterson, and H.~Ghasemzadeh, ``On-device machine learning for diagnosis of parkinson’s disease from hand drawn artifacts,'' in {\em 2022 IEEE-EMBS International Conference on Wearable and Implantable Body Sensor Networks (BSN)}, pp.~1--4, 2022.

\bibitem{wang2020convergence}
X.~Wang, Y.~Han, C.~Wang, Q.~Zhao, X.~Chen, and M.~Chen, ``Convergence of edge computing and deep learning: A comprehensive survey,'' {\em IEEE Communications Surveys \& Tutorials}, vol.~22, no.~2, pp.~869--904, 2020.

\bibitem{khan2020federated}
L.~U. Khan, S.~R. Pandey, N.~H. Tran, W.~Saad, Z.~Han, M.~N. Nguyen, and C.~S. Hong, ``Federated learning for edge networks: Resource optimization and incentive mechanism,'' {\em IEEE Communications Magazine}, vol.~58, no.~10, pp.~88--93, 2020.

\bibitem{mazilu2015prediction}
S.~Mazilu, A.~Calatroni, E.~Gazit, A.~Mirelman, J.~M. Hausdorff, and G.~Tr{\"o}ster, ``Prediction of freezing of gait in parkinson's from physiological wearables: an exploratory study,'' {\em IEEE journal of biomedical and health informatics}, vol.~19, no.~6, pp.~1843--1854, 2015.

\bibitem{luo2020hfl}
J.~Luo, B.~Song, M.~Wen, H.~Dong, and Y.~Zhang, ``Hfl: Hybrid federated learning for healthcare applications on edge devices,'' in {\em 2020 IEEE International Conference on Big Data (Big Data)}, pp.~3172--3181, IEEE, 2020.

\bibitem{li2020federated}
T.~Li, A.~K. Sahu, A.~Talwalkar, and V.~Smith, ``Federated learning: Challenges, methods, and future directions,'' {\em IEEE signal processing magazine}, vol.~37, no.~3, pp.~50--60, 2020.

\bibitem{rieke2020future}
N.~Rieke, J.~Hancox, W.~Li, F.~Milletari, H.~R. Roth, S.~Albarqouni, S.~Bakas, M.~N. Galtier, B.~A. Landman, K.~Maier-Hein, {\em et~al.}, ``The future of digital health with federated learning,'' {\em NPJ digital medicine}, vol.~3, no.~1, pp.~1--7, 2020.

\bibitem{zhao2018hybrid}
A.~Zhao, L.~Qi, J.~Li, J.~Dong, and H.~Yu, ``A hybrid spatio-temporal model for detection and severity rating of parkinson’s disease from gait data,'' {\em Neurocomputing}, vol.~315, pp.~1--8, 2018.

\bibitem{sama2012dyskinesia}
A.~Sama, C.~P{\'e}rez-L{\'o}pez, J.~Romagosa, D.~Rodriguez-Martin, A.~Catala, J.~Cabestany, D.~A. Perez-Martinez, and A.~Rodr{\'\i}guez-Molinero, ``Dyskinesia and motor state detection in parkinson's disease patients with a single movement sensor,'' in {\em 2012 Annual International Conference of the IEEE Engineering in Medicine and Biology Society}, pp.~1194--1197, IEEE, 2012.

\bibitem{naghavi2021towards}
N.~Naghavi and E.~Wade, ``Towards real-time prediction of freezing of gait in patients with parkinson’s disease: a novel deep one-class classifier,'' {\em IEEE Journal of Biomedical and Health Informatics}, vol.~26, no.~4, pp.~1726--1736, 2021.

\bibitem{koltermann2023fog}
K.~Koltermann, W.~Jung, G.~Blackwell, A.~Pinney, M.~Chen, L.~Cloud, I.~Pretzer-Aboff, and G.~Zhou, ``Fog-finder: Real-time freezing of gait detection and treatment,'' in {\em Proceedings of the 8th ACM/IEEE International Conference on Connected Health: Applications, Systems and Engineering Technologies}, pp.~22--33, 2023.

\bibitem{soumma2024self}
S.~B. Soumma, K.~Mangipudi, D.~Peterson, S.~Mehta, and H.~Ghasemzadeh, ``Self-supervised learning and opportunistic inference for continuous monitoring of freezing of gait in parkinson's disease,'' {\em arXiv preprint arXiv:2410.21326}, 2024.

\bibitem{xia2024prediction}
Y.~Xia, H.~Sun, B.~Zhang, Y.~Xu, and Q.~Ye, ``Prediction of freezing of gait based on self-supervised pretraining via contrastive learning,'' {\em Biomedical Signal Processing and Control}, vol.~89, p.~105765, 2024.

\bibitem{lonini2018wearable}
L.~Lonini, A.~Dai, N.~Shawen, T.~Simuni, C.~Poon, L.~Shimanovich, M.~Daeschler, R.~Ghaffari, J.~A. Rogers, and A.~Jayaraman, ``Wearable sensors for parkinson’s disease: which data are worth collecting for training symptom detection models,'' {\em NPJ digital medicine}, vol.~1, no.~1, p.~64, 2018.

\bibitem{chen2020fedhealth}
Y.~Chen, X.~Qin, J.~Wang, C.~Yu, and W.~Gao, ``Fedhealth: A federated transfer learning framework for wearable healthcare,'' {\em IEEE Intelligent Systems}, vol.~35, no.~4, pp.~83--93, 2020.

\bibitem{beaussart2021waffle}
M.~Beaussart, F.~Grimberg, M.-A. Hartley, and M.~Jaggi, ``Waffle: Weighted averaging for personalized federated learning,'' {\em arXiv preprint arXiv:2110.06978}, 2021.

\bibitem{tlvmc-parkinsons-freezing-gait-prediction}
A.~Howard, amit salomon, eran gazit, J.~Hausdorff, L.~Kirsch, Maggie, P.~Ginis, R.~Holbrook, and Y.~F. Karim, ``Parkinson's freezing of gait prediction.'' \url{https://kaggle.com/competitions/tlvmc-parkinsons-freezing-gait-prediction}, 2023.
\newblock Kaggle.

\bibitem{hammerla2016deep}
N.~Y. Hammerla, S.~Halloran, and T.~Pl{\"o}tz, ``Deep, convolutional, and recurrent models for human activity recognition using wearables,'' {\em arXiv preprint arXiv:1604.08880}, 2016.

\bibitem{naghavi2019prediction}
N.~Naghavi and E.~Wade, ``Prediction of freezing of gait in parkinson’s disease using statistical inference and lower--limb acceleration data,'' {\em IEEE transactions on neural systems and rehabilitation engineering}, vol.~27, no.~5, pp.~947--955, 2019.

\bibitem{yokkampon2022autoencoder}
U.~Yokkampon, A.~Mowshowitz, S.~Chumkamon, and E.~Hayashi, ``Autoencoder with gramian angular summation field for anomaly detection in multivariate time series data,'' {\em Journal of Advances in Artificial Life Robotics}, vol.~2, no.~4, pp.~206--210, 2022.

\bibitem{buz2020novel}
A.~C. Buz, M.~U. Demirezen, and U.~Yavano{\u{g}}lu, ``A novel approach and application of time series to image transformation methods on classification of underwater objects,'' {\em Gazi M{\"u}hendislik Bilimleri Dergisi}, vol.~7, no.~1, pp.~1--11, 2020.

\bibitem{mikos2017real}
V.~Mikos, C.-H. Heng, A.~Tay, N.~S.~Y. Chia, K.~M.~L. Koh, D.~M.~L. Tan, and W.~L. Au, ``Real-time patient adaptivity for freezing of gait classification through semi-supervised neural networks,'' in {\em 2017 16th IEEE International Conference on Machine Learning and Applications (ICMLA)}, pp.~871--876, IEEE, 2017.

\bibitem{10.1016/j.artmed.2022.102459}
L.~Borz\`{\i}, L.~Sigcha, D.~Rodr\'{\i}guez-Mart\'{\i}n, and G.~Olmo, ``Real-time detection of freezing of gait in parkinson’s disease using multi-head convolutional neural networks and a single inertial sensor,'' {\em Artif. Intell. Med.}, vol.~135, jan 2023.

\end{thebibliography}

\end{document}